\documentclass[11pt,a4paper]{article}
\usepackage[hyperref]{eacl2021}
\usepackage{times}
\usepackage{latexsym}
\usepackage{graphicx}

\usepackage{microtype}

\aclfinalcopy 
\def\aclpaperid{122} 

\title{AfriKI: Machine-in-the-Loop Afrikaans Poetry Generation}

\author{Imke van Heerden  \\
  Dept. of Comparative Literature \\
  College of Social Sciences and Humanities \\
  Koç University, Istanbul, Turkey \\
  \texttt{ivanheerden@ku.edu.tr} \\\And
  Anil Bas \\
  Dept. of Computer Engineering \\
  Faculty of Technology \\
  Marmara University, Istanbul, Turkey \\
  \texttt{anil.bas@marmara.edu.tr} \\}

\date{}

\begin{document}
\maketitle
\begin{abstract}
This paper proposes a generative language model called AfriKI. Our approach is based on an LSTM architecture trained on a small corpus of contemporary fiction. With the aim of promoting human creativity, we use the model as an authoring tool to explore machine-in-the-loop Afrikaans poetry generation. To our knowledge, this is the first study to attempt creative text generation in Afrikaans.
\end{abstract}

\section{Introduction}

Afrikaans\footnote{The Constitution of the Republic of South Africa recognises Afrikaans as one of eleven official languages, alongside Sepedi, Sesotho, Setswana, siSwati, Tshivenda, Xitsonga, English, isiNdebele, isiXhosa and isiZulu \citep{assembly1996constitution}. In South Africa, there are approximately 6.9 million first-language speakers of Afrikaans, according to the most recent census \citep{lehohla2012census}.} is a language spoken largely in South Africa, Namibia, Botswana and Zimbabwe. Masakhane \citep{nekoto2020participatory, orife2020masakhane} draws important attention to the current disproportion of NLP research and resources with respect to African languages. In fact, in the entire ACL Anthology,\footnote{https://www.aclweb.org/anthology/} of the thirteen studies that mention ``Afrikaans'' in their titles, only four \citep{sanby2016comparing, augustinus2016afribooms, dirix2017universal,ralethe2020adaptation} appeared in the last five years. By no means do we ignore studies with inclusive \citep{eiselen2014developing} and multilingual approaches \citep{ziering2016towards} or those published via other platforms \citep{van2003improving}. This is simply an indication that NLP research in Afrikaans is limited, especially in comparison to resource-rich languages, i.e. the so-called ``winners'' in the taxonomy of \citet{joshi2020state}.

In this paper, we present a generative language model called AfriKI, an abbreviation for ``Afrikaanse Kunsmatige Intelligensie'' (\emph{Afrikaans Artificial Intelligence}). We use this model as an authoring tool to explore machine-in-the-loop poetry generation in Afrikaans. Machine-in-the-loop frameworks promote human creativity through computational assistance, as opposed to human-in-the-loop pipelines, which aim to strengthen machine learning models \citep{clark2018creative}. We treat poetry generation as a hybrid system, an experimental approach that enables the generation of high-quality poetic text with very limited data. To our knowledge, this is the first study in creative text generation as well as an initial step towards automatic poetry generation in Afrikaans.

Whereas NLG in its quest for full automation may frown upon human involvement, our human-centred framework does the opposite. According to \citet{lubart2005can},

\vspace{-2mm}
\begin{quote}
one criticism of artificial intelligence programs that claim to be creative is exactly that a human plays a role at some point, which reduces the autonomy of the machine. From the HCI perspective [...] these ``failed'' AI creativity programs are examples of successful human–computer interactions to facilitate creativity.
\end{quote}
\vspace{-2mm}

This study demonstrates that human-machine collaboration could enhance human creativity. We agree with \citet{shneiderman2002creativity} that support tools ``make more people more creative more often''.

\section{Related Work}

Several computational models focus on automatic poetry generation. First approaches follow rule-based, template-based systems \citep{gervas2001expert,diaz2002poetry}. \citet{levy2001computational} and \citet{manurung2012using} apply genetic algorithms while \citet{jiang2008generating} and \citet{he2012generating} use statistical machine translation, with \citet{yan2013poet} utilising text summarisation to generate poetry. \citet{oliveira2009automatic} provides a clear overview of early systems and presents a comparable method \citeyearpar{oliveira2012poetryme}.

Starting with \citet{zhang2014chinese}, we have seen great advancements in poetry generation using neural networks. \citet{wang2016attention} extend this using the attention mechanism \citep{bahdanau2014neural}. There are many attempts to improve the quality of learning-based generated poetry, by using planning models \citep{wang2016chinese}, finite-state machinery \citep{ghazvininejad2016generating}, reinforcement learning \citep{yi2018automatic} as well as variational autoencoders \citep{yang2018generating}.

Conventional recurrent neural networks (RNN) are not suitable for learning long range dependencies \citep{wang2016attention} due to the vanishing gradient problem \citep{bengio1994learning}. Long short-term memory (LSTM) networks \citep{hochreiter1997long} address this issue and are widely used for language modeling \citep{sundermeyer2012lstm}. \citet{tikhonov2018sounds} propose word-based LSTM to generate poetry. \citet{potash2015ghostwriter} adopt a similar technique to produce rap lyrics. \citet{zugarini2019neural} apply syllable-based LSTM to generate tercets. Finally, composed of various LSTM models, Deep-speare \citep{lau2018deep} generates Shakespearean sonnets.

The remarkable quality and results of these studies are indisputable. However, they all concentrate on data-rich languages such as English, Chinese, Italian and Russian. For example, the character language model of \citet{hopkins2017automatically} uses a poetry corpus consisting of 7.56 million words and 34.34 million characters. Likewise, a recent study by \citet{liu2020deep} trained on over 200 thousand poems and 3 million ancient Chinese prose texts.

We trained an LSTM network for poetic text generation as well. However, our approach differs in significant ways. First, whereas these studies generate verse in a fully automatic manner, we emphasise human creativity, introducing a strong computational component to the creative writing process. Second, the aforementioned studies either trained on comprehensive poetry datasets or model poetic qualities. To illustrate the latter, the recent work of \citet{van2020automatic} focuses on specifically non-poetic text in English and French, however, is able to model the rhyme constraint using phonetic representation of words from Wiktionary. Since there is no publicly available large-scale poetry dataset in Afrikaans, we follow an alternative approach, constructing our model as a text generator that produces individual sentences and phrases instead of stanzas of verse. In other words, the model outputs a set of lines, which we arrange vertically into short poems without modification.

\section{Model}

In this section, we explain the dataset, model architecture as well as the co-creative poetry generation process.

\begin{figure}[!t]
\noindent\resizebox{.49\textwidth}{!}{
\includegraphics[height=5cm, clip]{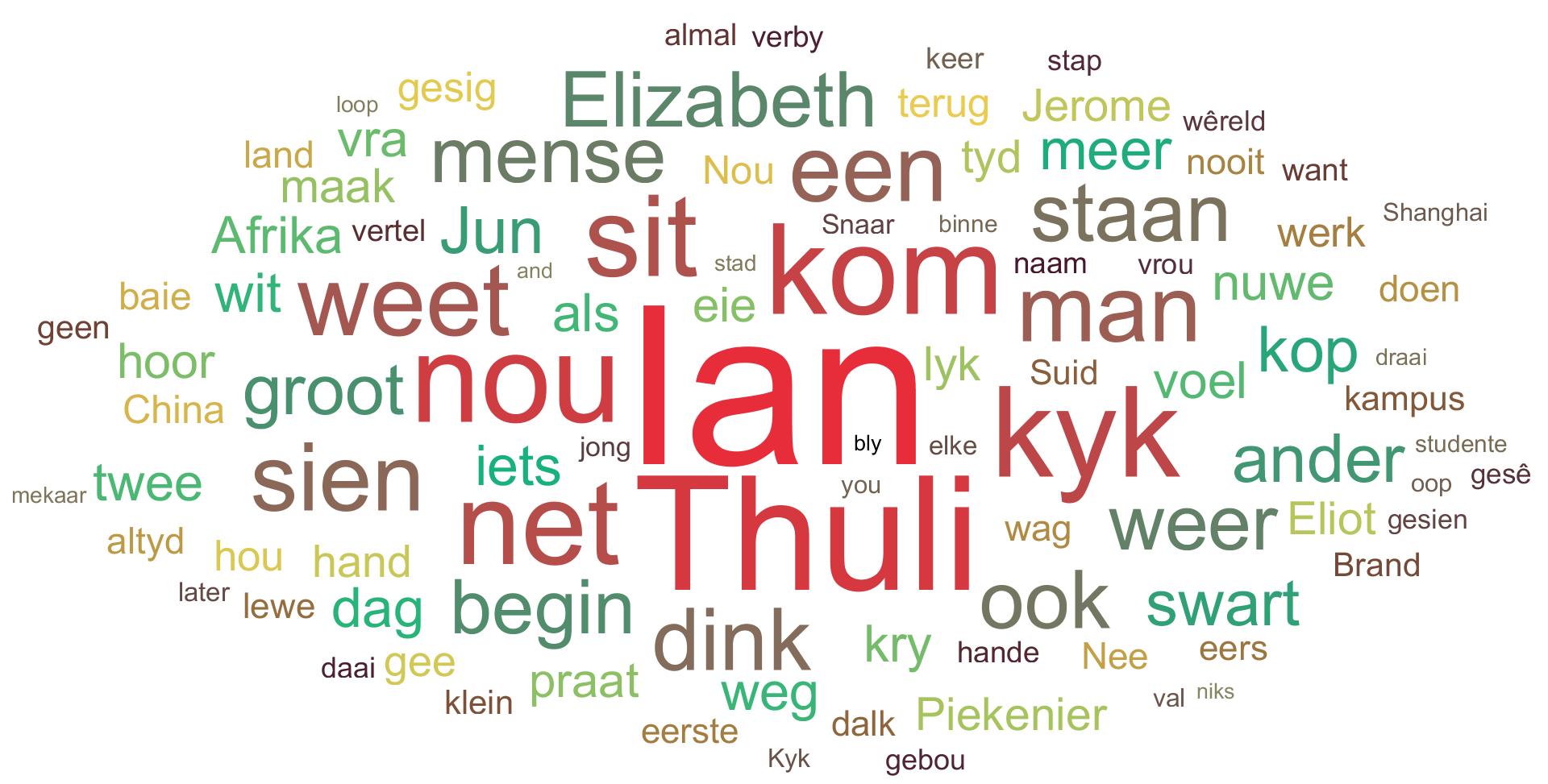}}
\caption{Frequently occurring words in \emph{Die Biblioteek aan die Einde van die Wêreld}. Stop words were removed. Note that Ian and Thuli are the protagonists.}
\label{fig:wordcloud}
\end{figure}

\paragraph{Corpus:} AfriKI trained on a lengthy (208,616-word) literary novel titled \emph{Die Biblioteek aan die Einde van die Wêreld} (\emph{The Library at the End of the World}) \citep{van2019biblioteek} by the South African novelist Etienne van Heerden. In 2020, the book was awarded the University of Johannesburg Prize for Literature \citep{pienaar2020book}. This work of new journalism combines fictional techniques with documentary language, and is particularly suitable given its use of rich imagery, figurative language as well as different Afrikaans varieties like \emph{Kaaps} (or Cape Afrikaans) and Standard Afrikaans. Figure~\ref{fig:wordcloud} shows a word cloud of its most commonly used words.

\begin{table*}[ht!]
\centering
\begin{tabular}{cc}
\hline
\textbf{Original (Afrikaans)} & \textbf{Translation (English)}\\
\hline
\\
\emph{Die konstabel se skiereiland} & \emph{The constable’s peninsula}  \\ \vspace{-2mm}
\\ 
Afrika drink                        & Africa drinks                     \\
onheil in die water.                & disaster in the water.            \\
Die landskap kantel sy rug          & The landscape tilts its back      \\
in sigbewaking en vlam.             & in surveillance and flame.        \\
Ons oopgesnyde sake                 & Our cut-open affairs              \\
brandtrappe vir die ander state.    & fire escapes for other states.    \\
Hierdie grond word intimidasie.     & This soil becomes intimidation.    \\ \vspace{-1mm}
\\
\hline
\\
\emph{Gedigte, daar by die brul van ’n brander} & \emph{Poetry, there near the roar of a wave}  \\ \vspace{-2mm}
\\
Hier is die oë katvoet vir                      & Here the eyes are cautious of                 \\
die spoelrotse onder uitdrukkings               & the sea rocks under expressions               \\
die golwe van gister wat                        & the waves of yesterday that                       \\
getol en woes en water                          & whirled and wild and water                    \\
saam met die son skuim in hul woorde            & froth with the sun in their words        \\ \vspace{-2mm}
\\ 
die ingedagte see                               & the introspective sea                         \\
lig die geure en praat                          & lifts the scents and utters                   \\
’n asemhaal                                     & a breath                                      \\ \vspace{-1mm}
\\
\hline
\\
\emph{Kaapstad}                     & \emph{Cape Town}                                  \\ \vspace{-2mm}
\\
Vandag is ons nie net die stad nie  & Today we are not just the city                    \\
maar                                & but                                               \\
die vertaler van die son            & the translator of the sun                         \\ \vspace{-2mm}
\\
Vanaand se gordyne                  & Tonight’s curtains                                \\
glinster by skuifvensters           & glitter at sliding windows                        \\
in die stadsliggies                 & in the city lights                                \\ \vspace{-2mm}
\\
Die uur van die winde               & The hour of the winds                             \\
sorg dat dit rondom klink          & takes care it sounds around                       \\
Sy wil die glasvensters deurkosyn   & She wants to doorframe the glass windows   \\
eens iets te beskerm                & to protect something                              \\ \vspace{-2mm}
\\
Tafelberg                           & Table Mountain                                    \\
maak ’n vraag waarbinne ons         & creates a question in which we                    \\
’n duisend name                     & are given                                         \\
genoem word                         & a thousand names                                  \\ \vspace{-1mm}
\\
\hline
\end{tabular}
\caption{\label{table:poetry-examples} Example results of machine-in-the-loop poetry generation.}
\end{table*}

\paragraph{Model Architecture:} Experimenting with several architectures, including LSTM, Multi-Layer LSTM and Bi-LSTM, we obtain best results with the following two-layer LSTM architecture. We use a vanilla LSTM structure \citep{hochreiter1997long} and, to avoid repetitiveness, omit to describe the network diagram and equations, similar to \citet{sundermeyer2012lstm}. We start with 100-dimensional word embeddings with a vocabulary size of 23,317 words, where weights are randomly initialised from a normal distribution with zero mean and standard deviation 0.01. Next, we stack two LSTM layers with 50 units in each layer followed by dropout layers with the rate of 0.2. This is followed by a fully connected layer and a softmax layer. We use the Adam optimiser \citep{kingma2015adam} with a learning rate = 0.001, batch size = 16, and train for 300 epochs. Although tweaking the parameters did change the model performance, it was not significant.

\paragraph{Machine-in-the-Loop:} Human-machine collaboration for the enhancement of creative writing has been examined under automated assistance \citep{roemmele2015creative,roemmele2018linguistic}, co-authorship \citep{tucker2019machine}, co-creativity \citep{manjavacas2017synthetic, kantosalo2019experience, calderwood2020novelists}, interactive storytelling
\citep{swanson2012say,brahman2020cue} and machine-in-the-loop \citep{clark2018creative, akoury2020storium}. 

Applying \citet{clark2018creative}'s terminology, we employ an iterative interaction structure that follows a push method of initiation with low intrusiveness. To clarify, our process consists of a single loop with two stages. First, the model generates a sizable set of unique individual lines (hundreds). Although memory networks may repeat parts of the training data \citep{ghazvininejad2016generating}, the generated phrases are highly distinct from the dataset, with hardly any repetition of word order. Second, the first author responds by choosing phrases at will. To create the final artefact, the author arranges the selected lines vertically. Generated text is used strictly without modification (except for some capitalisation and punctuation). The result of our collaborative writing system is short, compelling works of poetry that draw inspiration from the literary movements Imagism \citep{hughes1972imagism} and Surrealism \citep{balakian1986surrealism}.

\section{Results}

Table~\ref{table:poetry-examples} presents three examples of poems produced by means of the co-creative process. Here, we discuss quality from a literary perspective.

Trained on prose, the text is generated as free verse (i.e. free from the restrictions of rhythm and rhyme) which we associate with contemporary poetry. In the lines, various poetic devices can be identified, such as alliteration (e.g. ``\textbf{g}olwe van \textbf{g}ister'') and assonance (e.g. ``m\textbf{aa}k 'n vr\textbf{aa}g w\textbf{aa}rbinne'').

The generated lines abound with figurative language as well. As an instance of an extended metaphor, the first stanza of the second poem suggests sensitivity to the country's turbulent history. Personification is particularly prevalent, lending a visceral quality to the text: Africa drinks, the landscape tilts its back, the sea breathes, and Table Mountain poses a question. The imagery is vivid, portraying sight (\emph{Tonight’s curtains / glitter at sliding windows / in the city lights}), smell (\emph{the introspective sea / lifts the scents and utters / a breath}) and sound (\emph{roar of a wave}). The language can be described as minimalist, evocative and  abstract, and therefore open to interpretation, resembling Imagist and Surrealist poetry.

Afrikaans has a rich poetic tradition \citep{brink2000groot}, and we believe that creative text generation has the potential to enrich poetic language. Alongside Afrikaans varieties, the corpus contains some English as well, which influenced the generated text in interesting ways. As one example, it is grammatically incorrect in Standard Afrikaans to use ``sun'' as both noun and verb, e.g. ``to sun in the garden''. The model, however, adopted this and other patterns from the English, generating novel phrases (that do not sound anglicised) such as ``sonlig son die promenade'' – \emph{sunlight suns the promenade}.

\section{Conclusion}

In this study, we present Afrikaans poetry generation in a machine-in-the-loop setting. Each and every line of poetry is automatically generated by the proposed LSTM network. In order to clearly identify the machine's contribution to the process, the human writer’s interaction is limited to the selection and vertical arrangement of the lines -- without any modification. We believe this is the first creative text generation study in the Afrikaans language. More broadly, the work encourages human-centred design in low-resource languages. Creative industries would benefit from co-creative tools and methods \citep{hsu2019users}, perhaps more than fully automatic approaches.

\section{Future Work}

There are many ways in which this work can be extended.

First, similar to \citet{yi2017generating}, we could follow line-to-line poem generation, where the network takes the previous line as prompt and generates a new line which, in turn, is the prompt for the next entry. We could also experiment with different architectures, such as Transformer \citep{vaswani2017attention}, as well as training schemes. For example, we could borrow AfriBERT \citep{ralethe2020adaptation}, the recent BERT \citep{devlin2019bert} adaptation for Afrikaans, to apply transfer learning.

Second, as demonstrated in \citet{van2020automatic}, poetry generation is also possible by training on prosaic (non-poetic) text and modeling poetic constraints (e.g. rhyme). This way, we could expand to fully automatic poetry generation. Naturally, this would require an extensive literature corpus.

Third, regarding the unconventional use of some nouns as verbs in Afrikaans, future research could explore how prevalent this type of novel, cross-language variation is. To improve textual quality, we could incorporate Afrikaans datasets such as the NCHLT Annotated Text Corpora \citep{eiselen2014developing, Puttkammer2014NCHLT} as well as the Afrikaans treebank \citep{augustinus2016afribooms}, which are available via SADiLaR \citep{roux2016south} in addition to others.

Finally, a promising direction to pursue would be the involvement of poets and writers to investigate whether this approach could inform and improve their creative writing practices.

\section*{Acknowledgments}

This paper has been produced benefiting from the 2232 International Fellowship for Outstanding Researchers Program of TÜBİTAK (Project No: 118C285). However, the entire responsibility of the paper belongs to the owner of the paper. The financial support received from TÜBİTAK does not mean that the content of the publication is approved in a scientific sense by TÜBİTAK.

We would like to thank Etienne van Heerden for providing his manuscript to be used in this study.

\bibliography{eacl2021}
\bibliographystyle{acl_natbib}

\end{document}